\documentclass[11pt]{article}

\usepackage[final]{acl}

\usepackage{times}
\usepackage{latexsym}
\usepackage{booktabs}
\usepackage{url}

\usepackage{xcolor}

\usepackage{listings}
\usepackage[T1]{fontenc}

\usepackage[utf8]{inputenc}

\usepackage{microtype}

\usepackage{inconsolata}

\usepackage{graphicx}

\title{A Dunning–Kruger–Like Dissociation in LLM Evaluation}
\title{Scaling Effects in Likelihood- and Prompt-Based Evaluation of LLMs}
\title{Likelihood- and Prompt-Based Evaluation Mismatch in LLMs Across Scale}
\title{Likelihood Ranking doesn't Scale Like Prompting in LLMs}

\author{
 \textbf{Alessandro Bondielli\textsuperscript{1,2,*}},
 \textbf{Lucia Passaro\textsuperscript{1,2,*}},
 \textbf{Davide Bacciu\textsuperscript{2}},
 \textbf{Alessandro Lenci\textsuperscript{1}}
\\
\\
 \textsuperscript{1}CoLingLab, Department of Philology, Literature and Linguistics, University of Pisa\\
 \textsuperscript{2}Department of Computer Science, University of Pisa\\
\\
 \small{
   *Equal contribution. \textbf{Correspondence:} \href{mailto:alessandro.bondielli@unipi.it}{alessandro.bondielli@unipi.it}, \href{mailto:lucia.passaro@unipi.it}{lucia.passaro@unipi.it}}
}

\begin{document}
\maketitle
\begin{abstract}
LLM evaluation is commonly performed either by prompting models to produce answers or by scoring candidate outputs with likelihood-based metrics. In multiple-choice QA, however, standard likelihood-based scoring is still conditioned on the question and answer set, and can therefore leverage the same task-conditioned answer-selection interface used in prompting. 
We study a complementary protocol based on likelihood ranking of declarative statements constructed from the same question--answer pairs. Across 95 decoder-only models, ranging from 0.1B to 104B parameters, and 10 MCQA datasets, we find a systematic divergence between declarative-statement likelihood ranking and prompted answering. Statement-likelihood accuracy remains comparatively stable across scale, whereas prompted answering improves sharply with scale and instruction-tuning. These results suggest that likelihood preferences over controlled declarative alternatives and task-conditioned answer selection probe distinct aspects of model behavior, and should not be treated as interchangeable.

\end{abstract}

\section{Introduction}
Evaluation protocols play a central role in shaping conclusions about capabilities of LLMs. Two paradigms that dominate current practice are \textbf{prompting-based evaluation}, in which models are directly asked to produce answers, often using a multiple-choice format \cite{hendrycks2021measuring}, and \textbf{likelihood-based evaluation} \cite{hu-levy-2023-prompting}, in which candidate outputs are scored using perplexity or related metrics.

\begin{figure}[t]
    \centering
    \includegraphics[width=\columnwidth]{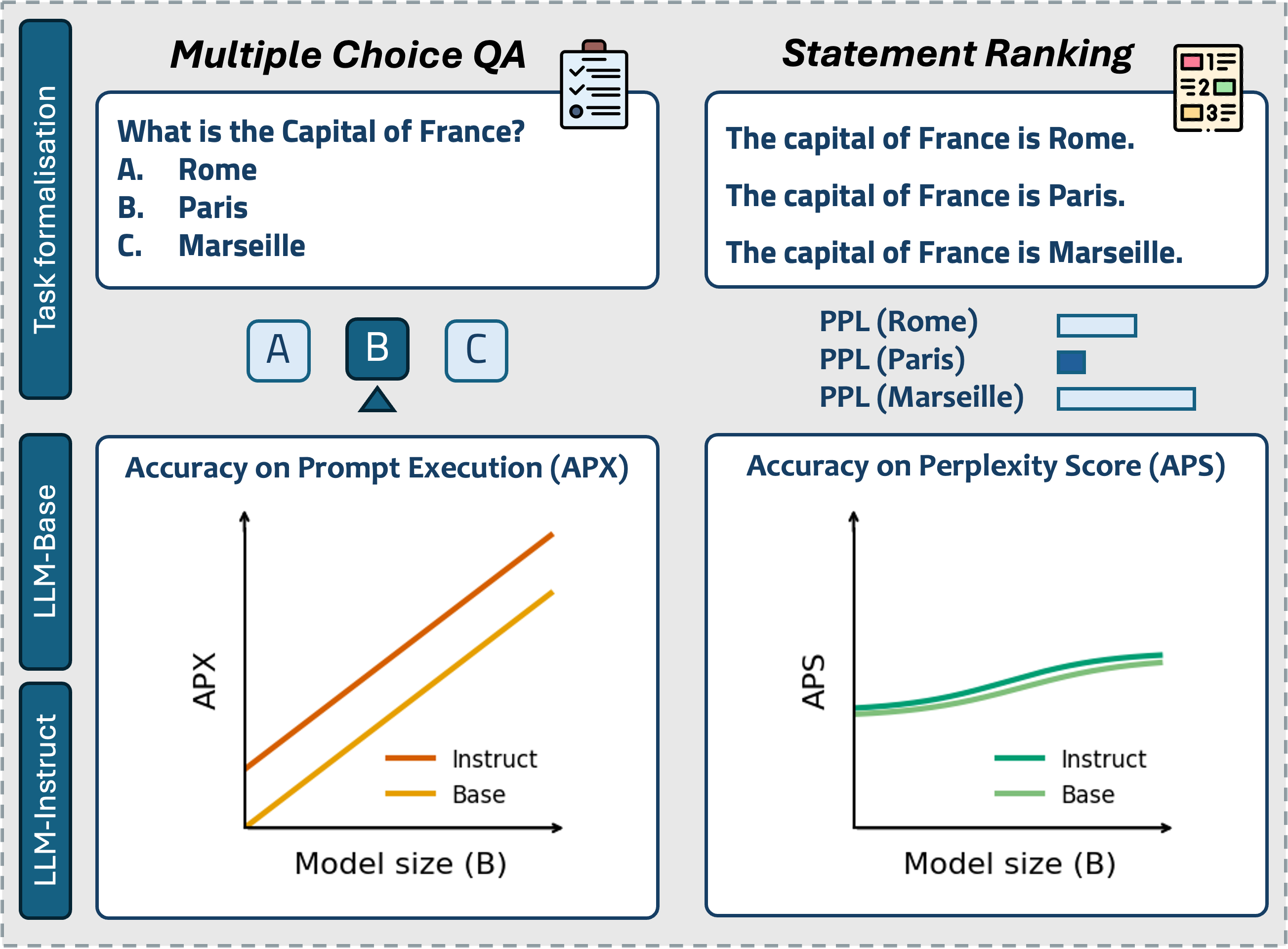}
    \caption{Overview of the experimental setting and evaluation. The design enables a comparison between prompting-based evaluation and declarative-statement likelihood ranking.}
    \label{fig:figure1}
\end{figure}

Prompting-based evaluation requires the model to map its likelihood preferences onto a discrete action conditioned on task framing, instructions, and output conventions. This mapping constitutes a learned behavioral policy that is shaped by model scale and alignment.
By contrast, Likelihood-based evaluation probes how probability mass is distributed over alternatives, e.g. factual statements, under a fixed linguistic form. In this work, we instantiate this paradigm through declarative statements derived from multiple-choice questions \cite{petroni-etal-2019-language}.

There is mounting evidence in the literature that prompting based evaluations, especially those involving multiple-choice selections, are inherently flawed and lack robustness \cite{wei2024rethinking,llm-mcq-bias,molfese-etal-2025-right,balepur-etal-2025-best}. Likelihood-based approaches are less ``task-oriented'', and have been shown to provide informative signals about linguistic and semantic plausibility \cite{hu-levy-2023-prompting,kauf-etal-2024-log}. However, likelihood scores are sensitive to formulation and surface form, making it important to distinguish different likelihood-based protocols. Standard likelihood-based MCQA still scores answer options in the original multiple-choice context; declarative-statement ranking removes this explicit answer-selection interface.
This distinction is related to recent work showing that multiple-choice answer selection may diverge from token-level likelihoods, and that likelihood estimates are affected by surface-form competition \cite{wang2024looktextinstructiontunedlanguage,wang-etal-2024-answer-c,holtzman-etal-2021-surface}. How these signals diverge with scale and instruction-tuning is underexplored.

Here, we present large scale empirical evidence that declarative-statement likelihood ranking and prompting-based evaluations systematically diverge as a function of model scale and post-training. Our main findings are: i.) likelihood-based accuracy over declarative statements improves slowly and remains relatively bounded across scale, while prompting-based accuracy exhibits strong scaling behavior and rapidly surpasses it; ii.) instruction-tuning accelerates the divergence, shifting the crossover point to smaller model sizes.

\section{Evaluation Setup}

As illustrated in Figure~\ref{fig:figure1}, we formulate two closely related tasks on the same data, namely \textbf{Multiple-Choice Question Answering} (MCQA) and \textbf{Statement Ranking}, to compare prompting-based evaluation with declarative-statement likelihood ranking. Model performance is assessed with accuracy under two complementary metrics. We define Accuracy on Prompt Execution (APX) as the proportion of correct answers produced in response to explicit prompts, and Accuracy on Perplexity Score (APS) as the proportion of cases where the model assigns higher likelihood to the correct declarative statement than to distractors.
Note that we distinguish APS from standard likelihood-based MCQA, where candidate answers are scored in the original question--option context. We argue that under our testing condition it follows the same APX-like scaling, with no significant differences.\footnote{We explicitly show this on preliminary experiments done on 51 models in Appendix~\ref{app:mcqa_likelihood}.}
This motivates declarative-statement ranking as a complementary protocol that reduces the task-conditioned answer-selection interface, rather than replacing standard likelihood-based MCQA. Intermediate option-conditioned formulations are valuable, but reintroduce part of the selection context that APS is designed to abstract away from.

We tested models on 10 HuggingFace MCQA datasets. First, we unified them into a common format: question, choices (labeled, e.g., "A", "B", etc.), correct label, and its index in the choices list. We preserved the original choices order when available, and randomized it otherwise.
Then, to allow for declarative-statement likelihood comparison under controlled linguistic forms, we automatically construct $n$ affirmative declarative statements for each question with $n$ answer options, one per option, using \texttt{gpt-oss-20b} \cite{agarwal2025gpt}. We provided the model with the question, choices, and instructions to generate declarative statements for each of the choices, one per line. We filtered out cases in which the model failed to provide exactly $n$  statements.\footnote{Appendix \ref{app:dataset} details the dataset creation (parameters, prompt, error handling, answer distribution, and validation).}
Table \ref{tab:dataset_stats} summarizes the final dataset composition and statistics.\footnote{We will release the final dataset 
via HuggingFace upon acceptance of the paper.} Since APS depends on the quality of the automatically constructed statements, we validate the generation pipeline for semantic fidelity and surface-form consistency.
On a stratified 1\% sample of the dataset, we observe a high similarity of generated statements to their source question--answer pair: BLEU mean/median is 0.69/0.75; BERTScore F1 mean/median is is 0.68/0.73; the median length gap is -1 token; the meaning preservation under two LLM judges (raw agreement 0.98) is 99\% . Within each item, a LLM judge finds candidate statements sharing a structural template in 95\% of cases. Full details on the validation process are provided in Appendix \ref{app:statement_validation}.


\begin{table}[t]
\centering
\setlength{\tabcolsep}{3pt}
\scriptsize
\begin{tabular}{p{3cm}cc}
\toprule
\textbf{Dataset (HF Name)} & \textbf{\# Question} & \textbf{Avg. Options ($\pm$ Std.)} \\
\midrule
TIGER-Lab/MMLU-Pro              & 10{,}632 & 9.45 $\pm$ 1.50 \\
allenai/ai2\_arc                & 7{,}698  & 4.00 $\pm$ 0.07 \\
allenai/openbookqa              & 11{,}836 & 4.00 $\pm$ 0.00 \\
allenai/qasc                    & 8{,}685  & 8.00 $\pm$ 0.00 \\
allenai/sciq                    & 13{,}627 & 4.00 $\pm$ 0.00 \\
maveriq/bigbenchhard            & 4{,}539  & 4.73 $\pm$ 3.09 \\
tau/commonsense\_qa             & 10{,}322 & 5.00 $\pm$ 0.00 \\
truthfulqa/truthful\_qa         & 714      & 4.92 $\pm$ 1.73 \\
yesilhealth/Health\_Benchmarks  & 7{,}358  & 4.00 $\pm$ 0.00 \\
yusuke1997/mCSQA                & 13{,}499 & 5.00 $\pm$ 0.00 \\
\midrule
Total                           & 88{,}910 & 5.35 $\pm$ 2.08\\
\bottomrule

\end{tabular}
\caption{Statistics of multiple-choice datasets used in our evaluation, same order as the table: \citet{mmlupro,arc,mihaylov-etal-2018-suit,KhotCGJS20,welbl-etal-2017-crowdsourcing,suzgun-etal-2023-challenging,talmor2019commonsenseqa,lin2022truthfulqa,yesilhealth_health_benchmarks,sakai-etal-2024-mcsqa}.
}
\label{tab:dataset_stats}
\end{table}

We tested a total of 95 open-weights decoder-only language models, ranging from around 0.1B to 104B parameters (see Table \ref{tab:model_families}, Appendix \ref{app:model_inventory}). 
To avoid confounding factors, we excluded Mixture-of-Expert and reasoning-enabled models from the evaluation. We consider both the instruction-tuned (henceforth, \textbf{instruct}) and pre-trained only (henceforth, \textbf{base}) variants, when available.\footnote{The full model inventory is available in Appendix \ref{app:model_inventory}.}

In the prompting-based setting, models were provided with instructions to solve the MCQA task, the MC question, its options, and prompted to directly and only provide the correct answer. We adapted the final part of the prompt to address differences between base and instruct models, e.g., ``the correct answer is: '' vs. ``what is the correct answer?'' respectively. We also followed the chat template of the model, when available. We set temperature to 0 for greedy generation. Finally, we parsed the models' responses via regular expressions to obtain a single, final answer, that we then compared with the ground truth to compute accuracy.
In the declarative-statement likelihood setting, we passed each declarative statement independently through the model, and computed its perplexity as $\exp\!\left(- \frac{1}{N} \sum_{i=1}^{N} \log p(x_i \mid x_{<i}) \right)$. Then, we selected the statement with the lowest perplexity and compared it with the ground truth to compute accuracy. All our experiments were conducted on HuggingFace models using vLLM.\footnote{See Appendix~\ref{app:eval_setup} for further details.}

\section{Results}

\begin{figure}[htbp]
    \centering
    \includegraphics[width=\columnwidth]{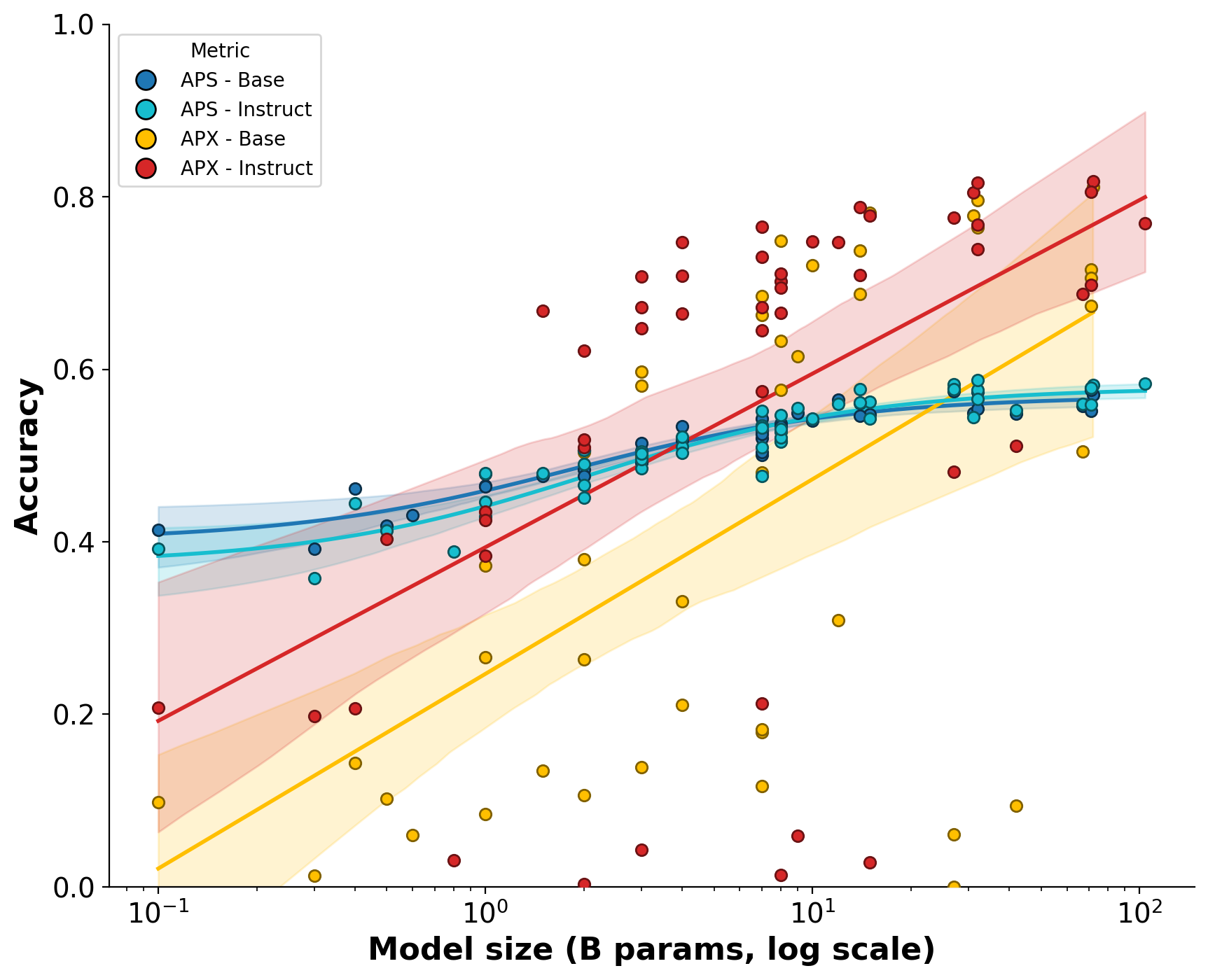}
\caption{Accuracies under prompting (APX) and declarative statement likelihood (APS) settings, across model scales and families. APS shows weak scaling and limited gains from instruction-tuning; APX improves sharply with scale, and larger models outperform APS.}
    \label{fig:apxvsaps}
\end{figure}

Figure \ref{fig:apxvsaps} illustrates the relationship between model scale and performance under APS and APX, across both base and instruct models (each point corresponds to an individual model variant).

To better characterize the relationship between scale and performance, we fit parametric scaling curves separately for each evaluation metric (APS, APX) and model type (base, instruct).
We fit several monotonic scaling functions and select the one minimizing the Akaike Information Criterion (AIC) \cite{Aikake-1974-statistical}, computed from the residual sum of squares. AIC enables the comparison of non-nested models while penalizing overparameterization.
The trend lines shown in Figure \ref{fig:apxvsaps} correspond to the best function fit for each subset.
Finally, we estimate uncertainty around each scaling curve via non-parametric bootstrap resampling \cite{efron1994bootstrap}. Shaded regions in the plot indicate 95\% confidence intervals.

Across datasets and model families, we observe a consistent pattern. APS exhibits weak scaling behavior: declarative-statement likelihood accuracy improves gradually with model size and remains within a relatively narrow range, saturating well below the best prompted results. Instruction-tuning has only a limited effect on this metric. Both base and instruct models follow roughly the same rational function, with minor differences particularly for smaller models. APX, by contrast, improves sharply with scale. Smaller models often perform substantially worse when prompted than when evaluated via declarative-statement likelihood ranking, while larger models exhibit rapid gains and ultimately outperform APS by a wide margin. We refer to the threshold at which APX surpasses APS as the \textit{prompting break-even point}. The best fit for both base and instruct models is a log-linear function. Both are very similar in terms of steepness, but instruction-tuned models' APX reaches the break-even point with APS substantially earlier. It is also worth noticing the much smaller variance of APS scores than APX ones, with the latter more loosely spread in the accuracy space. This pattern differs widely from the standard likelihood-based MCQA control reported in Appendix~\ref{app:mcqa_likelihood}, which instead follows similar scaling behavior as APX. This supports the interpretation that declarative-statement ranking captures a signal distinct from the task-conditioned answer-selection interface. We observe the same qualitative patterns by breaking down the analysis by dataset.\footnote{See Appendix \ref{app:additional_results} for Figures and further details.}

These trends indicate that prompted answer selection and declarative-statement likelihood ranking respond to different drivers of improvement. While likelihood preferences over declarative alternatives slowly improve with scale, the ability to act on these preferences under task prompts is strongly amplified by instruction-tuning. 
APX and APS exhibit similar trends across datasets, but the prompting break-even point varies: \textsc{Sciq} requires larger models than \textsc{OpenBookQA}; moreover APS on \textsc{Sciq}  is consistently high (0.65--0.85), suggesting that correct statements receive more stable likelihood support than in other datasets, with lower and more stable PPL.\footnote{See Appendix~\ref{app:additional_results} for further details.}

\section{Interpreting the mismatch}

Declarative-statement likelihood ranking probes how probability mass is distributed over the various alternatives under a fixed linguistic form. This signal is often diffuse, with small differences between correct and incorrect statements, and improves only gradually with scale. We verify this by computing the average PPL delta between the first and the second preferred answer. We see that, aside from few outliers, notably in the 4--10B parameter range, all models display near-zero differences between their first and second ``choice'' in terms of PPL.\footnote{See Appendix~\ref{app:additional_results} for details and visualizations.}

Prompting-based evaluation, by contrast, requires the model to interpret task instructions, compare alternatives, and commit to a discrete decision. This does not mean that prompted gains are disconnected from knowledge; rather, they conflate distributional support for the correct answer with the model's ability to express that support through the requested answer format. This behavior shows a learned answer-selection policy that maps diffuse likelihood preferences onto task-appropriate outputs. Improvements under this metric reflect not only sharper underlying distributions, but also more effective interfaces between distributional preferences and decision-making.

We view instruction-tuning as primarily optimizing the interface between likelihood preferences and task-conditioned behavior, rather than substantially reshaping the underlying probability distribution learned in pre-training. By training models to produce explicit, task-appropriate outputs under natural language instructions, instruction-tuning strengthens the mapping from diffuse distributional preferences to discrete decisions. The gap widens with scale: prompted accuracy increasingly reflects improved decision behavior and task compliance rather than proportional gains in declarative-statement likelihood accuracy. As model size, training data, and post-training procedures are often opaque in current model releases, we interpret the observed gap as reflecting the combined effect of scale and instruction-tuning on the answer-selection interface. Declarative-statement likelihood ranking and prompting-based evaluations target different objectives, with the latter resembling an easier discriminative decision task. As a result, gains in prompted performance may reflect improved task-conditioned decision behavior rather than sharper underlying distributions.
Our interpretation is related to work on factuality and calibration, including truth-evaluation protocols such as $P(\mathrm{True}\mid s)$ \cite{kadavath2022}. APS, however, is not intended as a calibrated truth estimator, but as a controlled likelihood-based signal for comparing declarative-statement ranking with prompted answer selection.
This signal remains sensitive to surface form, paraphrase competition, and candidate wording. 

Our validation mitigates these confounds by preserving the original question wording and enforcing structural similarity across candidate statements. Further normalizations, such as discounting unconditional option fluency, may help isolate statement-conditioned preferences more precisely. Nevertheless, our central finding is comparative: under the same controlled statement format, likelihood ranking and prompted answer selection scale differently across models, datasets, and instruction-tuning regimes.

\section{Conclusion}

Current evaluation practices risk conflating task accuracy with likelihood-based signals used to assess model capabilities. Instead, our findings suggest that these two practices capture different signals of model knowledge.
Improvements seen under prompting, particularly those induced by instruction-tuning, reflect advances in behavioral alignment and answer selection mechanisms at least as much as gains in likelihood preferences over semantically matched alternatives. 

We argue that prompting-based evaluation, standard likelihood-based MCQA, and declarative-statement likelihood ranking measure related but distinct properties and should not be used interchangeably. As LLMs continue to scale and are increasingly optimized for interactive use, evaluation protocols must more carefully distinguish between latent likelihood preferences and how effectively models are trained to act on them.

\section*{Limitations}

This study has several limitations.
First, we focus exclusively on MCQA. While this enables a controlled comparison between prompting-based and likelihood-based evaluation, it represents a limited class of tasks. 

Second, model families are not uniformly balanced across scale. Some families are overrepresented at particular sizes, which may introduce residual family-specific effects beyond scale.

Third, we consider only open-weights models. Although this supports reproducibility, it limits the generality of our findings to proprietary systems that may employ different alignment strategies.

Fourth, our analysis focuses on model scale and does not consider the amount of training tokens. Albeit the two are strongly correlated, exposure to different amounts of tokens could have independent effects and may be a confounding factor in the analysis. However, this aspect was not possible to analyze due to the lack of information on training data for many of the tested models, especially smaller variants of top-of-the-line models.

Fifth, APS relies on automatically generated declarative statements. Although we validate them for semantic fidelity, structural consistency, and length balance, residual surface-form differences, paraphrase competition, or stylistic preferences may still affect likelihood estimates \cite{holtzman-etal-2021-surface}. Thus, APS should be interpreted as a likelihood-based signal over controlled declarative alternatives, not as a direct or exhaustive measure of model knowledge.

Finally, our declarative-statement protocol is complementary to standard likelihood-based MCQA, where answer options are scored in the original multiple-choice context. We include this formulation as a control, but leave fuller comparisons with calibrated likelihood scores, option-fluency corrections, open-ended factual probes, and truth-evaluation protocols such as $P(\mathrm{True}\mid s)$ \cite{kadavath2022} to future work. We also leave a systematic qualitative analysis of APX--APS mismatch cases to future work.

Despite these limitations, the observed divergence between prompting-based and declarative-statement likelihood evaluation is robust across datasets and model families and highlights fundamental distinctions between evaluation paradigms.

\section*{Acknowledgments}

This work was supported by i) the PNRR MUR project \href{https://fondazione-fair.it/}{PE0000013-FAIR} (Spoke 1), funded by the European Commission under NextGeneration EU; ii) the EU EIC project \href{https://eic-emerge.eu}{EMERGE} (Grant No. 101070918); and iii) the PNRR MUR project FAIR TT\_02 ``Innovare la sorveglianza automatizzata delle infezioni del sito chirurgico tramite modelli di elaborazione del linguaggio naturale''.



\bibliography{custom}

\appendix


\section*{Appendix}

\section{Datasets Details}\label{app:dataset}

\subsection{Declarative Statements Generation}

To generate affirmative declarative statements for likelihood-based evaluation from multiple choice questions, we use \texttt{gpt-oss-20b}. We use the HuggingFace implementation with vLLM. Below we report the prompt used.

\paragraph{System prompt}
\begin{quote}
\footnotesize
\ttfamily
You are a precise text transformation assistant.
Your task is to generate, for each multiple-choice question, one affirmative
statement per answer option, both for correct and incorrect options.

If the question includes blanks (e.g., ``\_\_\_ is the capital of France.''),
fill them with each answer choice.

If not, form a natural affirmative statement by appending the answer, e.g.,
Question: ``What type of water formation is formed by clouds?''
Expected output format: ``The type of water formation formed by clouds is
[answer].''

Always output only the list of generated statements, one per line, in the same
order as the provided choices.

Each statement must:
\begin{itemize}
  \item Retain as much as possible the original text. Do not shorten or omit anything.
  \item Be rephrased only as needed to form a grammatically correct affirmative sentence.
  \item Insert or append the answer option in a grammatically coherent and correct way.
  \item Preserve the order of the answer options.
  \item Produce one output per choice, in the same order as the provided list.
  \item Be affirmative only (no question marks).
  \item Contain no commentary, explanations, or metadata.
\end{itemize}
\end{quote}

\paragraph{User prompt}
\begin{quote}
\footnotesize
\ttfamily
Question:\\
\{\{ question \}\}

\vspace{0.5em}
Choices:\\
\{\% for label, text in choices \%\}\\
\{\{ label \}\}. \{\{ text \}\}\\
\{\% endfor \%\}

\vspace{0.5em}
Generate one affirmative statement for each choice, maintaining their order and
preserving the question wording.
\end{quote}

Note that we also wrap the prompt in the original default \texttt{gpt-oss-20b} chat template,\footnote{\url{https://huggingface.co/openai/gpt-oss-20b/blob/main/chat_template.jinja}} which includes information on knowledge cutoffs and reasoning effort (see below).

\paragraph{Generation parameters}
We set the reasoning effort to ``medium'', $temperature=0.7$, and $top\_p=0.95$. We let the model generate a maximum of 1500 tokens, to fit both the reasoning trace and the final answer.

\begin{figure}[htbp]
    \centering
    \includegraphics[width=\columnwidth]{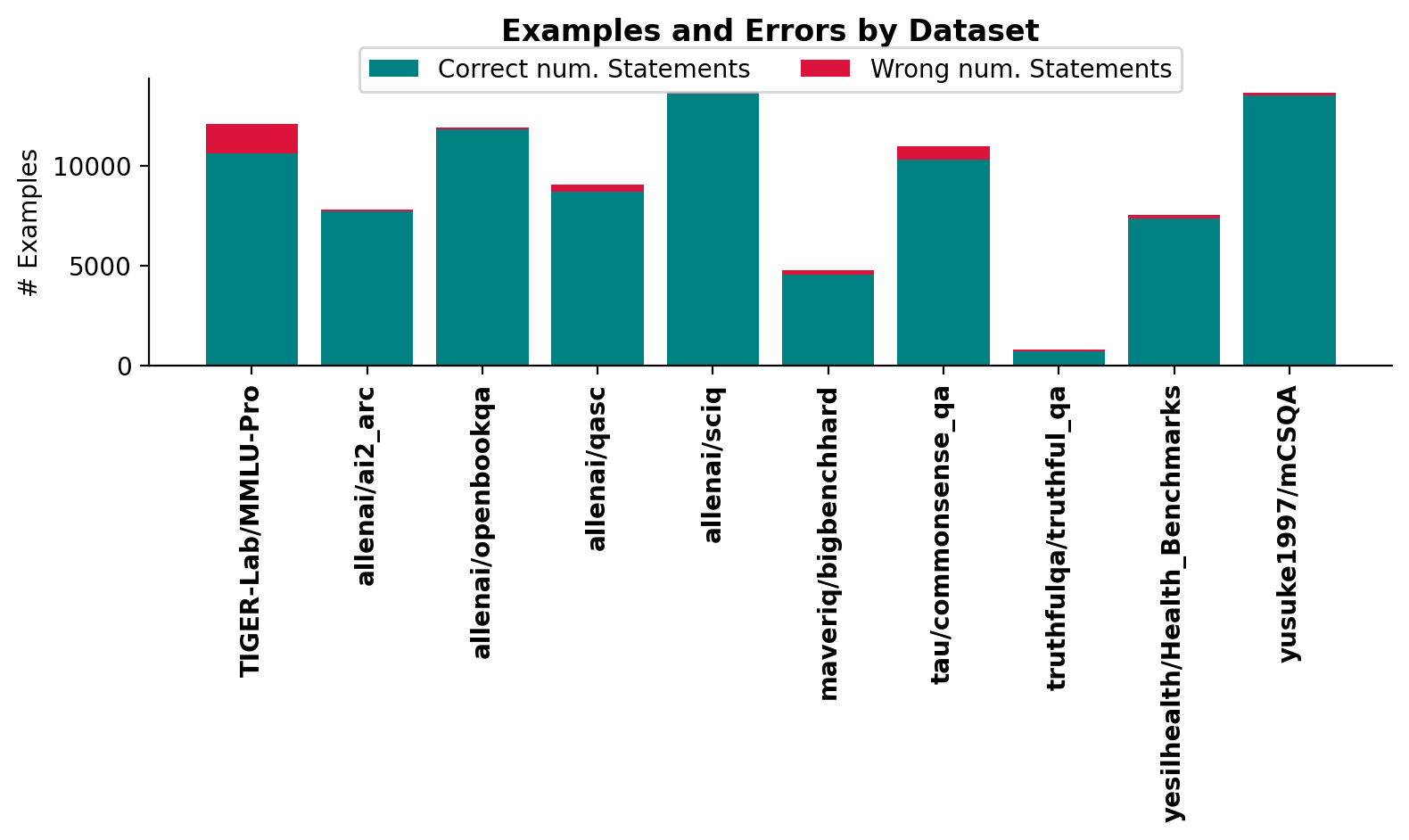}
\caption{Number of examples and errors in declarative statements generation for each dataset.}
    \label{fig:example-errors-dataset}
\end{figure}

\paragraph{Error Handling}
To  handle errors in generating declarative statements, we proceed as follows. We simply parse the model's response, excluding the reasoning trace, by splitting on new lines. Then, we remove empty strings and programmatically assert whether the number of declarative statements actually corresponds to the number of possible answers for that question. We filter cases where there were more or less declarative statements than possible answers. Figure \ref{fig:example-errors-dataset} shows the proportion of errors for each dataset. 

We see that the number of errors for each dataset is relatively small on all datasets except \textsc{MMLUPro}. This may be attributable to the fact that it is one of the most complex datasets, with the highest number of average choices (9.45$\pm$1.5).

\subsection{Statement Validation}\label{app:statement_validation}
We validate the generated declarative statements for semantic fidelity and surface-form consistency. 

\textbf{Semantic fidelity.} We sample 900 items, corresponding to approximately 1\% of the dataset, stratified by dataset and length bin to gather a representative sample of the dataset. 
We leverage LLM-as-a-judge with two independent judges, namely \texttt{gpt-oss-120b} (implemented locally, via vLLM) and \texttt{gpt-5.2} (via API). Judges are given the original question, answer option, and generated statement, and asked whether the statement preserves the meaning of the corresponding question--answer pair without omissions or distortions. The prompt used for both models is the following:
\begin{quote}
\footnotesize
\ttfamily
\textbf{[System]: }You are a strict semantic equivalence checker for multiple-choice QA.
 
Task: Determine whether the variable "statement" is semantically equivalent
to the variable "choice\_text" as an answer to the variable "question".

Context definition:
\begin{itemize}
    \item "question" is asked by Speaker A.
    \item Speaker B answers using exactly "choice\_text".
    \item "statement" is a candidate full sentence that may express B's answer.
\end{itemize}
 
Equivalence criteria:
\begin{itemize}
    \item The statement must assert exactly the same answer as choice\_text.
    \item Ignore superficial differences such as punctuation, capitalization, or grammatical form.
    \item Do not use external knowledge. Only compare semantic content.
    \item If the statement expresses a different answer, adds incompatible meaning, introduces uncertainty, or changes the asserted answer, it is not equivalent.\\
 \end{itemize}
Output format (JSON only, no extra text):
 
\{
  "equivalent": boolean,
  "choice\_label": string
\}

\vspace{0.5em}
\textbf{[User]: }Variables:\\
question: \{\{question\}\}\\
choice\_label: \{\{choice\_label\}\}\\
choice\_text: \{\{choice\_text\}\}\\
statement: \{\{statement\}\}\\
 
Determine whether "statement" is semantically equivalent to "choice\_text" as an answer to "question".
 
Return JSON only.
\end{quote}

Both judges label 99\% of the sampled statements as semantically equivalent, with raw agreement of 0.98. We do not report Cohen's $\kappa$ due to the strong class imbalance.

\begin{figure}[htbp]
    \centering
    \includegraphics[width=\columnwidth]{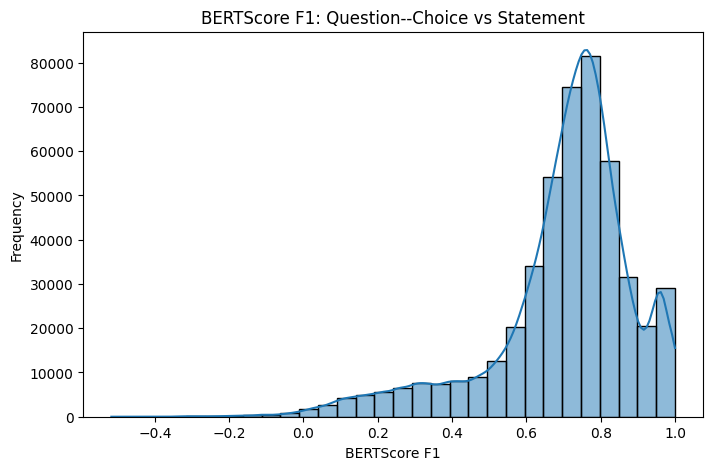}
\caption{Distribution of BERTScore F1 for each statement and question--answer pair.}
    \label{fig:bertscore-distribution}
\end{figure}

\begin{figure}[htbp]
    \centering
    \includegraphics[width=\columnwidth]{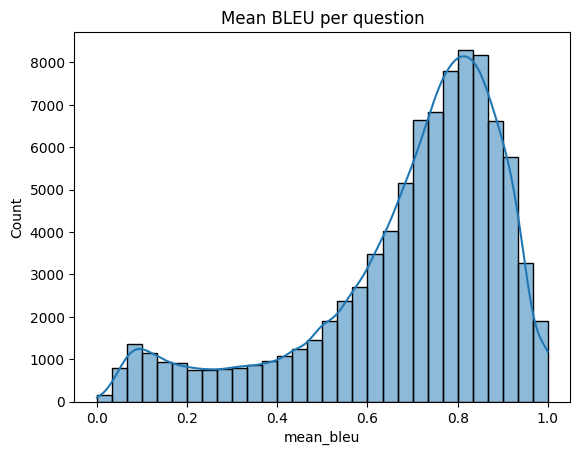}
\caption{Distribution of Mean BLEU scores per question.}
    \label{fig:mean-bleu}
\end{figure}

We further compute automatic similarity measures. We use BERTScore\footnote{Model: \texttt{roberta-large}} and BLEU\footnote{\texttt{sentence\_bleu}, NLTK implementation} to assess generated statements against their source question--answer pair. The BERTScore F1 mean/median is 0.68/0.73. Figure \ref{fig:bertscore-distribution} shows the complete distribution of F1 scores. The BLEU score is similar as well, with 0.69/0.75 mean/median across questions (Figure \ref{fig:mean-bleu}). 
Finally, we also assess wether candidate statements from the same item are comparable in length with their respective question--answer pair. We tokenize with the \texttt{gpt-oss-20b} tokenizer and obtain an average within-item (statement vs question--answer) median difference of -1.01 (mean -9.1),  indicating that the median statement is roughly 1 token shorter than its source question–-answer pair.

\textbf{Surface-form consistency. }
Finally, we perform an additional assessment of the structural consistency of generated statements, to verify whether generated statements for each question follow the same surface-form template. On the same 1\% sample, we ask a judge (\texttt{gpt-oss-120b}, implemented locally with vLLM) to determine whether all generated statements followed the same surface-form template. We propmt the model as follows:
\begin{quote}
\footnotesize
\ttfamily
\textbf{[System]: }You are an expert linguistic analyst evaluating structural and surface-form consistency across statements derived from the same multiple-choice question.

Your task is NOT to evaluate correctness.
Your task is NOT to evaluate semantics.
Your task is NOT to determine whether the answer is valid.
Your task is ONLY to determine whether the statements share the same surface-form template.

Two statements share a template if:
\begin{itemize}
    \item They have the same clause structure
    \item Negation appears in the same position
    \item Predicate framing is identical
    \item Tense and modality are identical
    \item Differences are limited to insertion of the answer content
\end{itemize}

You must:
\begin{enumerate}
    \item Look for a shared template (if present).
    \item Compare each statement to that template.
    \item Decide whether all statements follow the shared template.
\end{enumerate}

Return ONLY valid JSON.

Output format (JSON):\\
\{
"shared\_template": true or false
\}

\textbf{[User]: }Question:\\
\{\{ question \}\}

Original options:\\
\{\{ answers \}\}

Generated statements:\\
\{\{ statements \}\}

Instructions:
\begin{enumerate}
    \item Infer the surface-form template used by the statements.
    \item Replace the option-specific span with \texttt{[OPTION]}.
    \item Compare each statement to the template.
    \item Decide whether all statements share the same template.
\end{enumerate}

Return only JSON in exactly this format:

\{
  "shared\_template": true or false
\}

\end{quote}

The judge declared that 95\% of items have structurally consistent statements across answer options.

Despite limited by the size of the dataset sample, these results highlight that the generated dataset is viable for our evaluation.

\subsection{MCQA Answer Distribution}\label{app:answer-distribution}

\begin{figure}[htbp]
    \centering
    \includegraphics[width=\columnwidth]{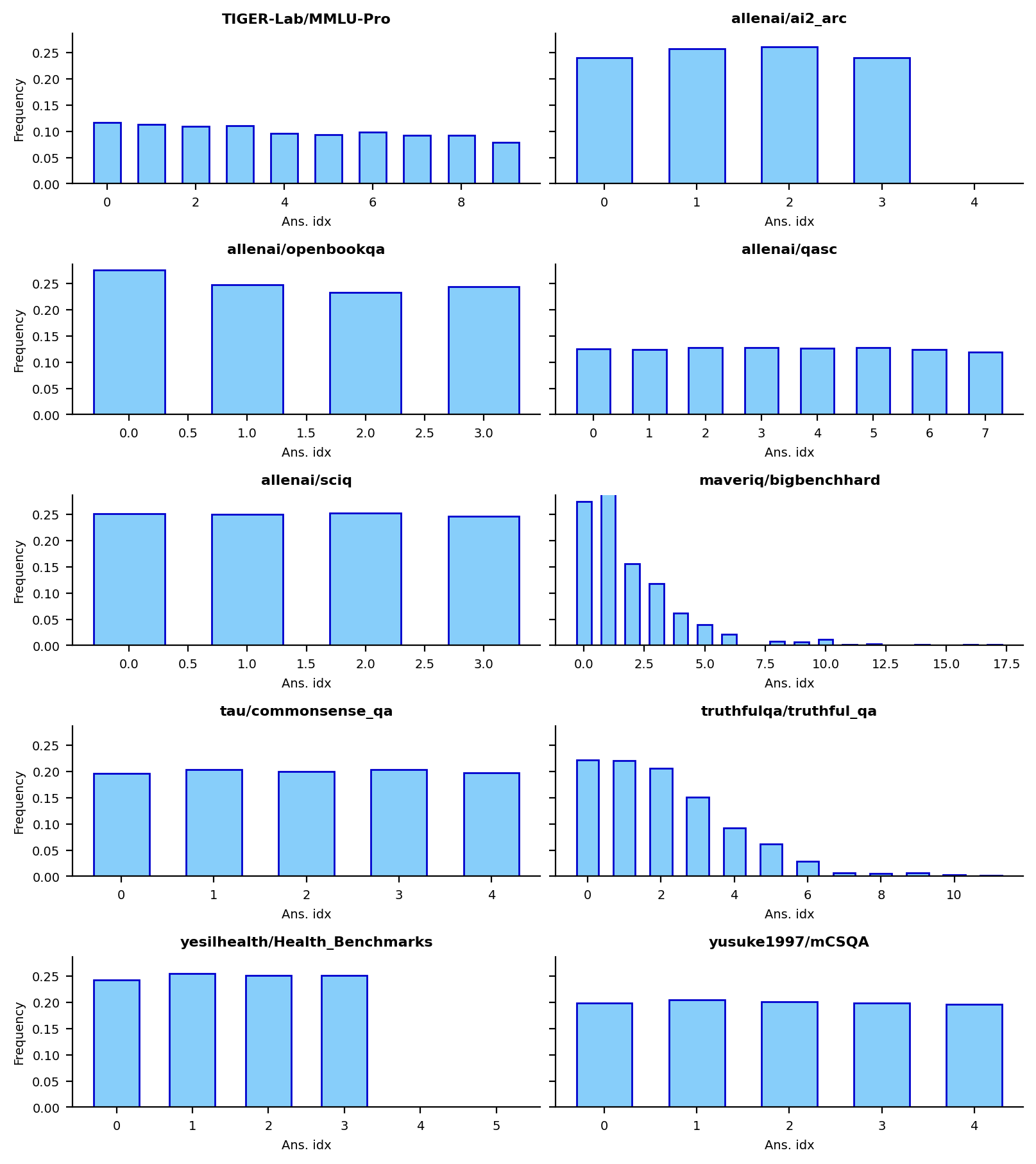}
\caption{Distribution of answers in each dataset.}
    \label{fig:answer-distribution}
\end{figure}

Figure \ref{fig:answer-distribution} shows the distribution of answers in each dataset. We observe that the distribution is uniform in most datasets. The main exceptions are \textsc{BigBenchHard} and \textsc{Truthful\_QA}. However, in both cases the distribution of answers is also attributable to a high variability in the number of possible answers for each question. If we look at Table \ref{tab:dataset_stats} we see in fact that both dataset have a much higher standard deviation than all other datasets.

\subsection{Dataset Licensing and Release}
All datasets are publicly available on HuggingFace and used in accordance with their licenses. We release our dataset\footnote{Via HuggingFace upon acceptance} including original QA pairs and generated declarative statements—under CC-BY, the most restrictive license among the sources.

\section{Standard MCQA Likelihood}\label{app:mcqa_likelihood}

In this Section, we aim to clarify the relation and highlight the difference between our formulation of APS and standard likelihood-based MCQA \cite{hu-levy-2023-prompting}. We clarify that our current formulation stems from preliminary experiments conducted early on a subset of models (i.e., $51$). These experiments act also as a control experiment in which we follow the common likelihood-based MCQA protocol. Specifically, each answer option is scored via perplexity as a continuation conditioned on the question and the full answer set, i.e., $\log P(\mathrm{option} \mid \mathrm{question}, \mathrm{options})$. We note that in these experiments we kept the prompt identical for both the base and instruct models. The prompt formulation is the one used for base models in our main experiments (see Appendix \ref{prompts_base_instruct}). This was done to limit variations and directly study APX vs likelihood-MCQA regardless of model tuning.

\begin{figure}[htbp]
    \centering
    \includegraphics[width=\columnwidth]{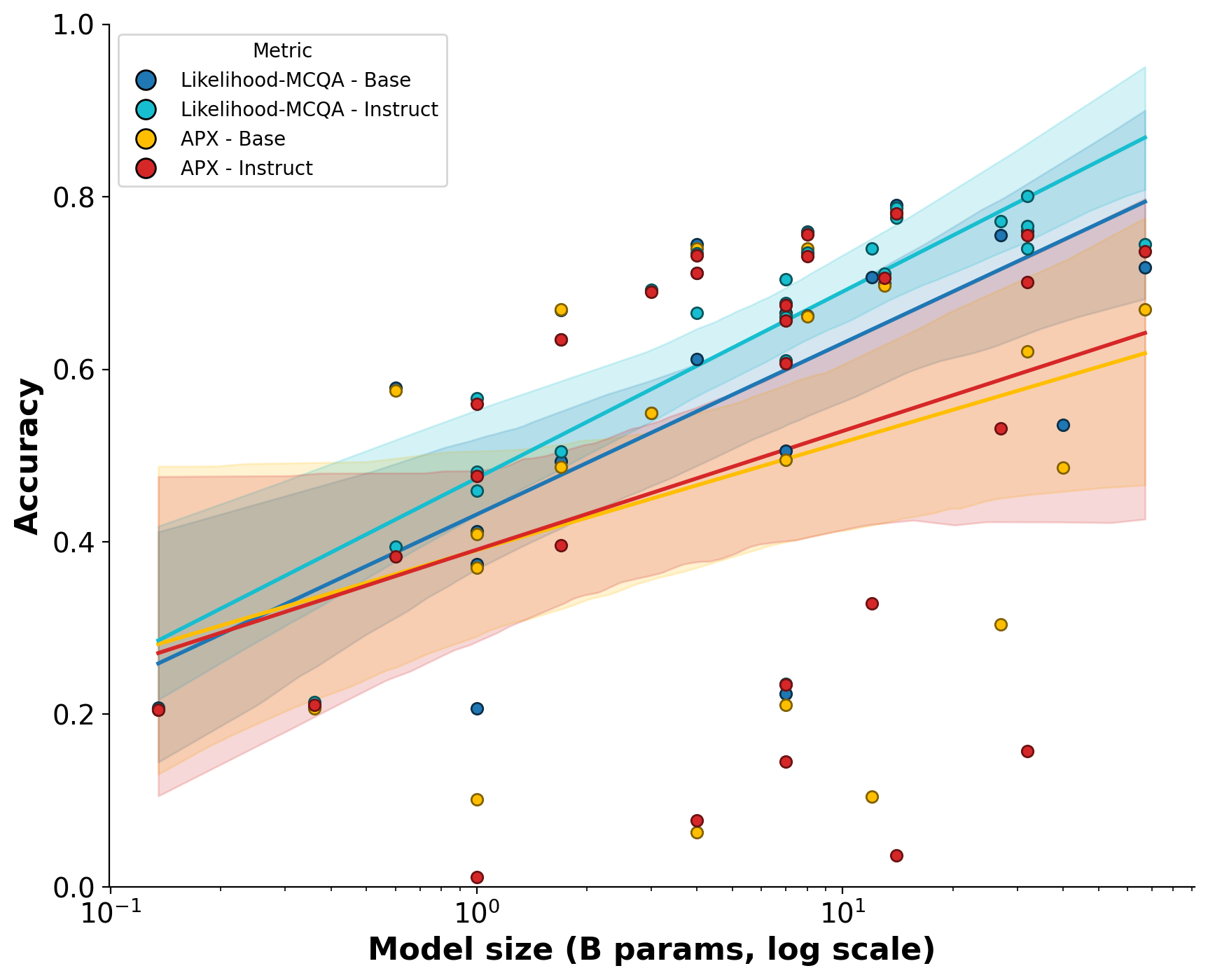}
\caption{Prompting-based (APX) and standard likelihood-based MCQA (Likelihood-MCQA) accuracy across model scales and families. Regardless of the presence of Instruction fine-tuning, likelihood-based MCQA and APX move in the same general direction, with comparable steepness.}
    \label{fig:L-MCQAvsAPX}
\end{figure}

With this setting we evaluate 51 models in the 0.135B--67B parameter range.
We provide a visualization analogous to that of our main experiment in Figure \ref{fig:L-MCQAvsAPX}. We plot performances against size, and we fit a log linear function to each of the groups.
This standard formulation of the problem produced trends highly similar to prompted answering. APX and standard option-likelihood scores are moderately correlated ($\rho=0.5444$, $p=3.025{\times}10^{-5}$). Moreover, a linear model with interaction terms shows no significant difference between the scaling slopes of APX and standard option-likelihood MCQA ($p=0.58$). This supports our interpretation that standard likelihood-based MCQA largely reflects the same task-conditioned answer-selection interface as greedy prompted answering, rather than an independent likelihood signal.

\section{Model Inventory}
\label{app:model_inventory}

Table \ref{tab:model_families} provides an overview of the model families evaluated in this work. Below we report the full list of evaluated models, grouped by family.

\begin{table}[!h]
\centering
\scriptsize
\begin{tabular}{lcccc}
\toprule
Family & \# Models & Base & Instruct & Size Range (B) \\
\midrule
Qwen2.5   & 14 & 7 & 7 & 0.5--72 \\
Qwen3     & 12 & 6 & 6 & 0.6--31 \\
Llama-3   & 9  & 5 & 4 & 1--71 \\
Gemma-3   & 10 & 5 & 5 & 0.3--27 \\
Gemma-2   & 6  & 3 & 3 & 2--27 \\
OLMo-2    & 8  & 4 & 4 & 1--32 \\
Falcon-3  & 8  & 4 & 4 & 1--10 \\
Falcon    & 4  & 2 & 2 & 7--42 \\
DeepSeek  & 4  & 2 & 2 & 7--67 \\
SmolLM2   & 6  & 3 & 3 & 0.1--2 \\
SmolLM3   & 2  & 1 & 1 & 3 \\
Apertus   & 4  & 2 & 2 & 8--71 \\
Command-R & 3  & 0 & 3 & 8--104 \\
Mistral   & 3  & 1 & 2 & 7--8 \\
Phi-4     & 2  & 0 & 2 & 4--15 \\
\bottomrule
\end{tabular}
\caption{Overview of the model families evaluated in this work, reporting the number of base and instruction-tuned variants and their parameter scale.}
\label{tab:model_families}
\end{table}

\paragraph{Phi-4.}
phi-4, Phi-4-mini-instruct.

\paragraph{Qwen2.5.}
Qwen2.5-0.5B, Qwen2.5-0.5B-Instruct, Qwen2.5-1.5B, Qwen2.5-1.5B-Instruct,
Qwen2.5-3B, Qwen2.5-3B-Instruct, Qwen2.5-7B, Qwen2.5-7B-Instruct,
Qwen2.5-14B, Qwen2.5-14B-Instruct, Qwen2.5-32B, Qwen2.5-32B-Instruct,
Qwen2.5-72B, Qwen2.5-72B-Instruct.

\paragraph{Qwen3.}
Qwen3-0.6B-Base, Qwen3-0.6B, Qwen3-1.7B-Base, Qwen3-1.7B,
Qwen3-4B-Base, Qwen3-4B-Instruct-2507, Qwen3-8B-Base, Qwen3-8B,
Qwen3-14B-Base, Qwen3-14B, Qwen3-30B-A3B-Base,
Qwen3-30B-A3B-Instruct-2507.

\paragraph{Llama-3.}
Llama-3.2-1B, Llama-3.2-1B-Instruct, Llama-3.2-3B, Llama-3.2-3B-Instruct,
Llama-3.1-8B, Llama-3.1-8B-Instruct, Llama-3.1-70B,
Llama-3.3-70B-Instruct, Meta-Llama-3-70B.

\paragraph{Gemma-2.}
gemma-2-2b, gemma-2-2b-it, gemma-2-9b, gemma-2-9b-it,
gemma-2-27b, gemma-2-27b-it.

\paragraph{Gemma-3.}
gemma-3-270m, gemma-3-270m-it, gemma-3-1b-pt, gemma-3-1b-it,
gemma-3-4b-pt, gemma-3-4b-it, gemma-3-12b-pt, gemma-3-12b-it,
gemma-3-27b-pt, gemma-3-27b-it.

\paragraph{OLMo-2.}
OLMo-2-0425-1B, OLMo-2-0425-1B-Instruct, OLMo-2-1124-7B,
OLMo-2-1124-7B-Instruct, OLMo-2-1124-13B,
OLMo-2-1124-13B-Instruct, OLMo-2-0325-32B,
OLMo-2-0325-32B-Instruct.

\paragraph{DeepSeek.}
deepseek-llm-7b-base, deepseek-llm-7b-chat,
deepseek-llm-67b-base, deepseek-llm-67b-chat.

\paragraph{Falcon-3.}
Falcon3-1B-Base, Falcon3-1B-Instruct, Falcon3-3B-Base,
Falcon3-3B-Instruct, Falcon3-7B-Base, Falcon3-7B-Instruct,
Falcon3-10B-Base, Falcon3-10B-Instruct.

\paragraph{Falcon.}
falcon-7b, falcon-7b-instruct, falcon-40b, falcon-40b-instruct.

\paragraph{SmolLM2.}
SmolLM2-135M, SmolLM2-135M-Instruct, SmolLM2-360M,
SmolLM2-360M-Instruct, SmolLM2-1.7B, SmolLM2-1.7B-Instruct.

\paragraph{SmolLM3.}
SmolLM3-3B, SmolLM3-3B-Base.

\paragraph{Apertus.}
Apertus-8B-2509, Apertus-8B-Instruct-2509,
Apertus-70B-2509, Apertus-70B-Instruct-2509.

\paragraph{Command-R.}
c4ai-command-r7b-12-2024, c4ai-command-r-08-2024,
c4ai-command-r-plus-08-2024.

\paragraph{Mistral.}
Mistral-7B-v0.3, Mistral-7B-Instruct-v0.3,
Ministral-8B-Instruct-2410.

\section{Details of Evaluation Setup}\label{app:eval_setup}

Here, we provide further details on the evaluation setup, including prompts used for MCQA for base and instruction-tuned models, and implementation via vLLM.

\subsection{Prompts}\label{prompts_base_instruct}

For the sake of comparison, we chose not to experiment with prompting techniques tailored to each specific model. We followed general prompting guidelines and kept the prompt short, simple and to the point.
We divided the prompt in two parts, namely \textbf{System} and \textbf{Question} prompts.
The System prompt is identical for all models. The main differences, that we report below, are in the way the question is asked to each model, with instruction-tuned models receiving a direct question, and base models a statement to complete, and in the fact that, when available, the prompt was wrapped into the chat template of each model. If the chat template was not available for the specific model (i.e., for base models and older instruction-tuned ones), we simply concatenated the System and Question prompts and fed them to the model.

\paragraph{System} 
\begin{quote}
\footnotesize
\ttfamily
You are an expert AI. Your task is to read a multiple-choice question and provide the most likely Correct Answer based on the Options given. 
Always output only the letter of the correct option. 
Do not add the actual answer, commentary, explanations, or metadata — only output the letter corresponding to the correct answer.
\end{quote}

\paragraph{Question - Instruct}
\begin{quote}
\footnotesize
\ttfamily
Question: \{\{ question \}\}

Options:\\
\{\% for option in options -\%\}\\
\{\{ option \}\}\\
\{\% endfor \%\}\\

Which is the correct answer? 
\end{quote}

\paragraph{Question - Base}
\begin{quote}
\footnotesize
\ttfamily
Question: \{\{ question \}\}
Options:\\
\{\% for option in options -\%\}\\
\{\{ option \}\}\\
\{\% endfor \%\}\\

The letter corresponding to the correct answer is: 
\end{quote}

Note that neither the instruct nor base variant have trailing whitespace after the final character. During early experimentation we found that while instruction-tuned variants were resilient to this kind of variation, several base models struggled if a trailing whitespace was added after the colon, often generating the end of sequence token. Thus, we chose to not include any trailing white space in the prompts.

\paragraph{Chat Templates}
For models that had it available, we wrap the prompt into the model's chat template as follows:

\begin{footnotesize}
\begin{verbatim}
messages = [
    {"role": "system", "content": SYSTEM},
    {"role": "user", "content": QUESTION},
]
\end{verbatim}
\end{footnotesize}

\subsection{Implementation}

All models were evaluated locally on a GPU node equipped with A100 80GB GPUs. Depending on the size of the model being tested, either one, two, or four GPUs were allocated for the experiment. For example, models smaller than 20B parameters could be fitted on a single GPU, while 70-100B parameter models required at least four GPUs to run. 
All models were evaluated using FP-16 variants provided by the original authors via HuggingFace.
Models were called using vLLM, specifically wrapping them into the \texttt{LLM} object. 

For MCQA, we simply let the model generate a maximum of 15 new tokens with temperature set to zero. To obtain a single, clear choice from model generations, we adopt a simple regex-based strategy where we search for possible answers in the generated text. We consider as possible answers either the letter in isolation, or the actual text of the answers. For cases in which the model provided more than a single answer, e.g., by repeating the original list of answers, we mark it as an error.

For likelihood-based evaluation, we pass each statement through the same pipeline, without any further prompting. We generate only one new token with temperature zero, and output log probabilities for the prompt. Then, we compute log probabilities of the whole sequence, and for each data point we assign a rank to the statements, from most likely (i.e., lowest logprobs) to the least likely.

\section{Additional Results}\label{app:additional_results}

\subsection{Per-dataset results}

\begin{figure}[t]
    \centering
    \includegraphics[width=.96\columnwidth]{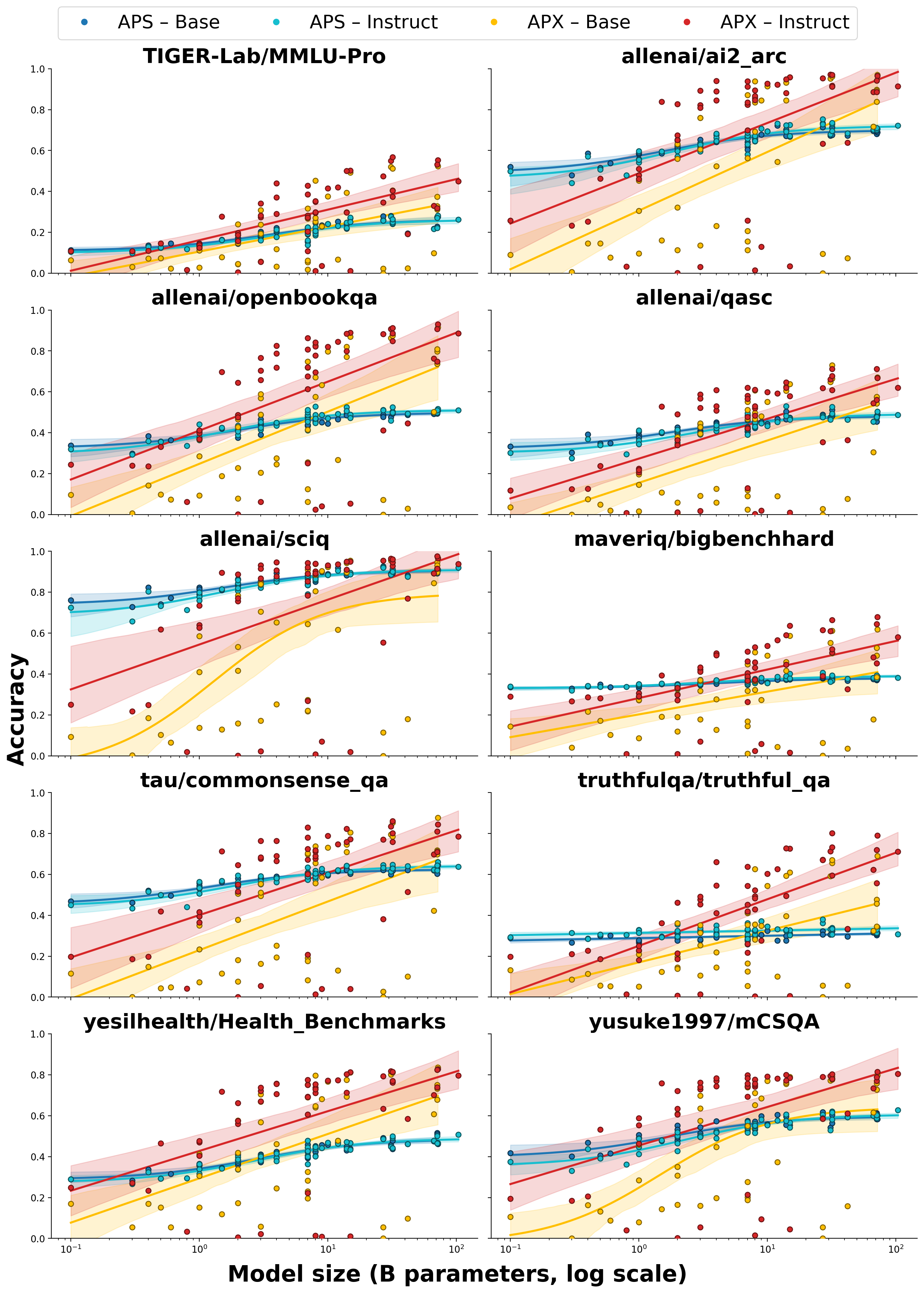}
\caption{Prompting-based and declarative-statement likelihood accuracies, for instruction-tuned and base models, as a function of model scale, for each dataset.}
    \label{fig:apxvsaps_bydataset}
\end{figure}

Figure~\ref{fig:apxvsaps_bydataset} reports the same analysis as Figure~\ref{fig:apxvsaps} broken down by dataset. We still observe the same qualitative pattern: APS exhibits weak scaling and limited sensitivity to instruction-tuning, while APX scales sharply with model size and benefits substantially from instruction-tuning. Although absolute performance levels vary across datasets, the divergent behavior between APS and APX is generally consistent, indicating that the mismatch is not driven by any single benchmark.

\subsection{PPL of correct answer}
\begin{figure}[htbp]
    \centering
    \includegraphics[width=0.8\columnwidth]{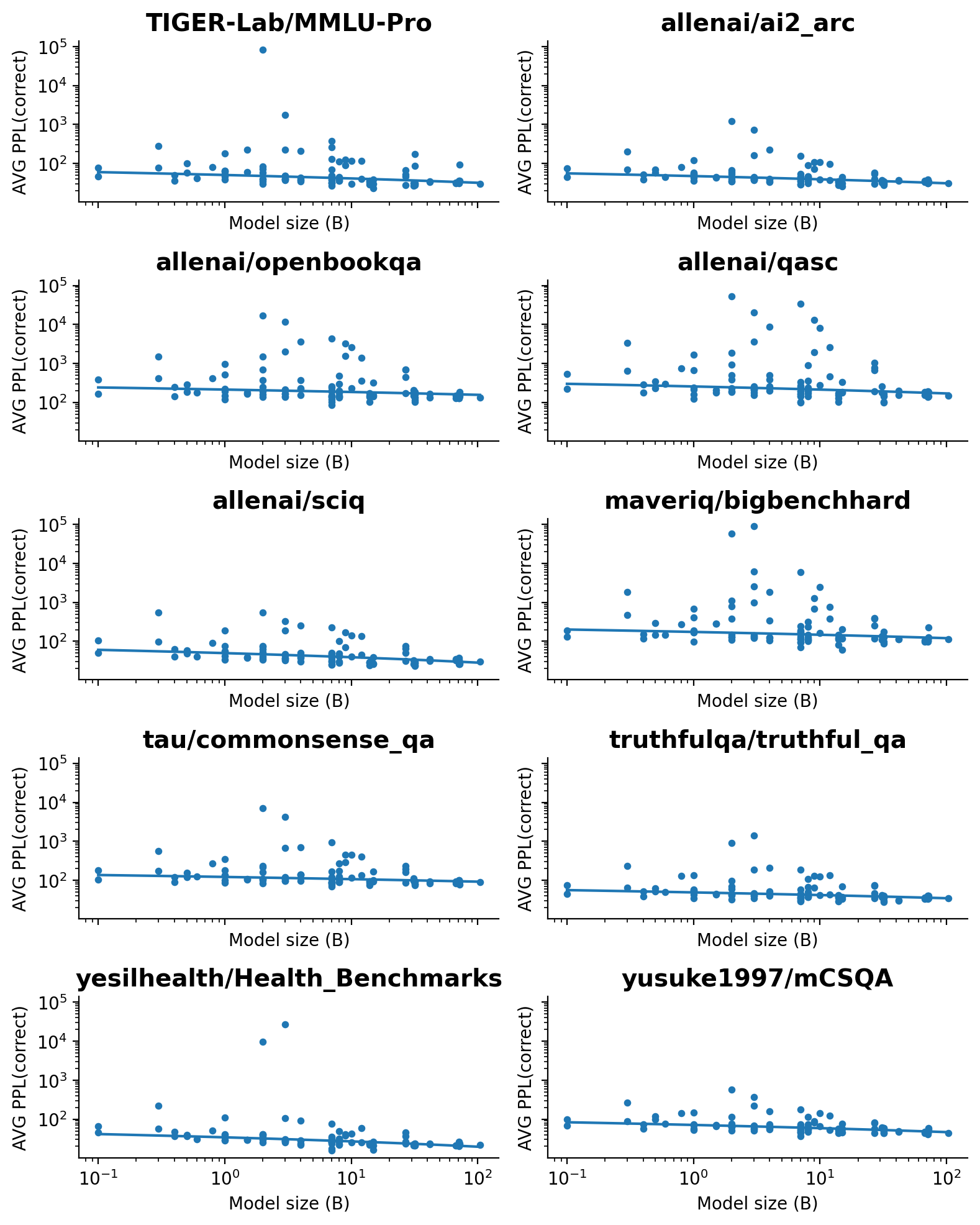}
\caption{Average PPL score assigned to the correct answer by each model, divided by dataset.}
    \label{fig:ppl-correct}
\end{figure}

Figure \ref{fig:ppl-correct} displays the average PPL score assigned to the correct answer by each model, divided by dataset. We observe that average PPL tends to decrease with model size, despite variability especially for middle-sized models. 

\subsection{PPL delta}
\begin{figure}[htbp]
    \centering
    \includegraphics[width=.8\columnwidth]{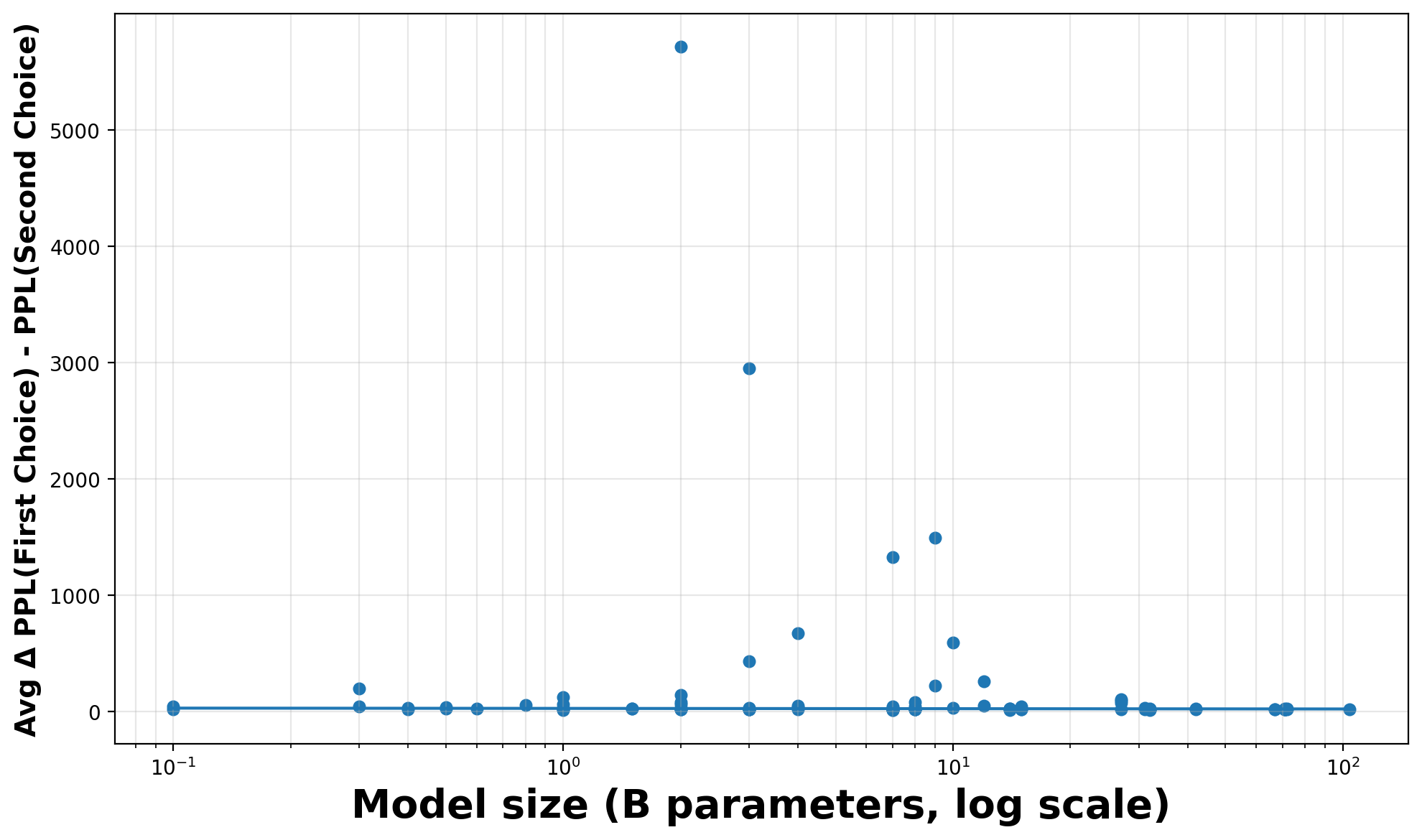}
\caption{Average delta between second lowest and lowest PPL score for declarative statements vs model size.}
    \label{fig:ppl-delta}
\end{figure}

Figure \ref{fig:ppl-delta} shows the delta between second lowest and lowest PPL score for declarative statements vs model size, i.e., between the model's actual choice and its second one. Aside from some notable excepitions, especially in the 4 to 10B parameter range, the difference between the models' choices and their second option is mostly near zero.

\end{document}